\PassOptionsToPackage{x11names,table}{xcolor}
\documentclass[10pt,twocolumn,letterpaper]{article}

\usepackage[pagenumbers]{iccv} % To force page numbers, e.g. for an arXiv version

\usepackage[normalem]{ulem}
\usepackage[x11names, table]{xcolor}
\usepackage{pifont}
\usepackage{multirow}
\usepackage{bbding}
\usepackage{adjustbox}
\usepackage[accsupp]{axessibility}  % Improves PDF readability for those with disabilities.
\newcommand{\tablestyle}[2]{\setlength{\tabcolsep}{#1}\renewcommand{\arraystretch}{#2}\centering\footnotesize}

\definecolor{iccvblue}{rgb}{0.21,0.49,0.74}
\definecolor{mygray}{gray}{.9}
\usepackage[pagebackref,breaklinks,colorlinks,allcolors=iccvblue]{hyperref}

\newcommand{\cmark}{\textcolor{green}{\ding{51}}} 
\newcommand{\xmark}{\textcolor{red}{\ding{55}}}  

\def\paperID{4009} % *** Enter the Paper ID here
\def\confName{ICCV}
\def\confYear{2025}

\title{\textit{\textcolor{violet}{Dis}\textcolor{DodgerBlue2}{Co}}: Towards \textcolor{violet}{Dis}tinct and \textcolor{DodgerBlue2}{Co}herent Visual Encapsulation in Video MLLMs}

\author{
    Jiahe Zhao$^1$, Rongkun Zheng$^2$, Yi Wang$^{3,4}$, Helin Wang$^5$, Hengshuang Zhao$^2$ \\
    \small{$^1$University of Chinese Academy of Sciences, $^2$The University of Hong Kong, $^3$Shanghai Artificial Intelligence Laboratory,} \\
    \small{$^4$Shanghai Innovation Institute, $^5$Fudan University} \\
    \tt\small{zhaojiahe22@mails.ucas.ac.cn, \{zrk22@connect, hszhao@cs\}.hku.hk,} \\
    \tt\small{wangyi@pjlab.org.cn, hlwang215@outlook.com}
}

\begin{document}
\maketitle
\begin{abstract}
In video Multimodal Large Language Models (video MLLMs), the visual encapsulation process plays a pivotal role in converting video contents into representative tokens for LLM input. While linear projectors are widely employed for encapsulation, they introduce semantic indistinctness and temporal incoherence when applied to videos. Conversely, the structure of resamplers shows promise in tackling these challenges, but an effective solution remains unexplored. Drawing inspiration from resampler structures, we introduce \textbf{DisCo}, a novel visual encapsulation method designed to yield semantically \textbf{dis}tinct and temporally \textbf{co}herent visual tokens for video MLLMs. DisCo integrates two key components: (1) A Visual Concept Discriminator (VCD) module, assigning unique semantics for visual tokens by associating them in pair with discriminative concepts in the video. (2) A Temporal Focus Calibrator (TFC) module, ensuring consistent temporal focus of visual tokens to video elements across every video frame. Through extensive experiments on multiple video MLLM frameworks, we demonstrate that DisCo remarkably outperforms previous state-of-the-art methods across a variety of video understanding benchmarks, while also achieving higher token efficiency thanks to the reduction of semantic indistinctness. The codes will be available at \href{https://github.com/ZJHTerry18/DisCo}{https://github.com/ZJHTerry18/DisCo}.  
\end{abstract}    
\section{Introduction}
\label{sec:intro}

% Background on video MLLMs
%% surge of MLLM -> First, image MLLM -> Then, Video MLLM -> Some popular video MLLMs
Multi-modal Large Language Models (MLLMs)~\cite{alayrac2022flamingo, li2023blip, driess2023palm, bai2023qwen, ye2024mplug, liu2024improved, chen2023sharegpt4v} have spearheaded the advancement of vision-language learning, gaining impressive visual understanding abilities on a myriad of open-world tasks. While the early exploitations were made on image inputs, recent studies have yielded profound breakthroughs on empowering MLLMs for video understanding~\cite{wang2024internvideo2, li2024mvbench, maaz2024videogpt+, chen2024sharegpt4video, korbar2025text, qian2024streaming, zheng2024syncvis}, contributing to a multitude of real-world applications like robotics~\cite{mu2024embodiedgpt}, autonomous driving~\cite{xu2024drivegpt4}, and AIGC~\cite{chen2024anydoor}. In contrast to images, video data is characterized by a substantially larger volume of visual information, coupled with inherent temporal complexities, which presents a formidable challenge in effectively encapsulating video inputs to facilitate optimal comprehension by the language model.

\begin{figure}[t]
    \centering
    \includegraphics[width=0.95\linewidth]{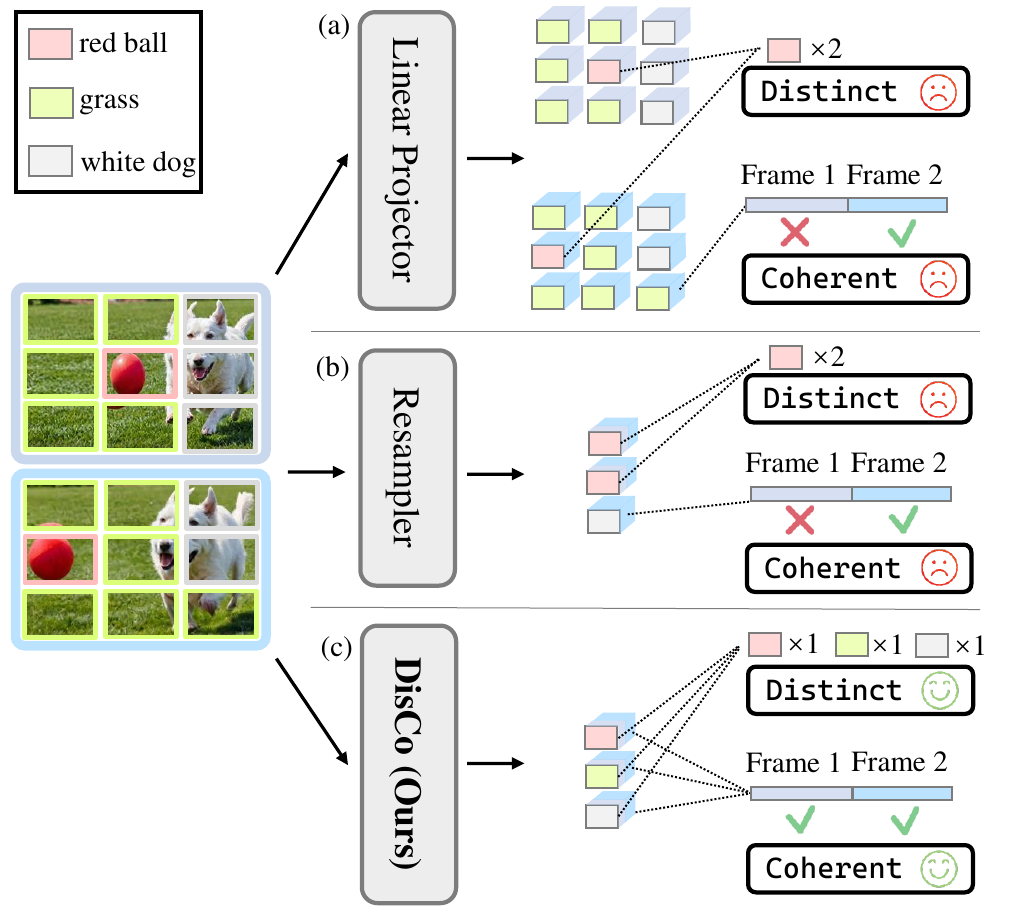}
    \caption{Illustrations of different visual encapsulation methods in video MLLMs. (a) \textit{Linear projector} directly projects tokens of each frame, leading to repetitive semantics for objects appearing in multiple frames, and is incapable of modeling cross-frame temporal coherence. (b) \textit{Resampler} utilizes attention mechanism to derive tokens, which is prone to redundant extraction of same semantics, and cannot guarantee coherent attention across frames. (c) Our proposed \textit{Disco} can generate high-quality tokens with distinct semantics and coherent temporal correlations.}
    \label{fig:teaser}
\end{figure}

% Current studies on the visual projectors of video MLLMs (Why we research on Q-Former)
For MLLMs, the visual connector~\cite{li2023blip, ge2023planting, cha2024honeybee, li2024tokenpacker, zhu2023minigpt} emerges as a pivotal component for the encapsulation of visual features into working tokens for LLM. Currently, a major stream of video MLLMs adopt linear projectors~\cite{liu2024visual, lin2023video, xu2024pllava, li2024topa} for visual encapsulation. While linear projector proficiently upholds local visual details with simple designs, as shown in~\cref{fig:teaser}(a), applying it to videos usually compromises performance since it introduces semantic indistinctness and temporal incoherence when processing videos. Specifically, the presence of repetitive visual elements across frames leads to redundancy in the projected tokens' semantics. Moreover, by discretely projecting each visual patch, linear projection fails to encapsulate temporal coherence across frames. More recently, a line of works~\cite{li2024tokenpacker, cha2024honeybee, chu2024mobilevlm} enhance the information compactness of linear projectors by downsampling or compressing the video patches. Nevertheless, the problems of semantic indistinctness and temporal incoherence are still not relieved due to the locality of their projection mechanisms in both spatial and temporal dimensions.

Different from linear projectors, resamplers~\cite{li2023blip, li2024mvbench, zhang2023video, korbar2025text, you2023ferret} exploit cross-attention that transforms video patches into a fixed set of visual tokens. This reduces indistinctness in form and implicitly conducts temporal modeling. However, we note semantic indistinctness and temporal incoherence still exist in the current resampler design. As depicted in~\cref{fig:teaser}(b), in resamplers, there are multiple tokens redundantly focusing on the same semantic instance, while neglecting other crucial instances. Meanwhile, visual tokens display poor temporal coherence by only attending to instances in a part of video frames while neglecting them in other frames. Intuitively, to enable LLMs to accurately comprehend video content, it is crucial to generate high-quality visual tokens representing diverse and distinct semantic concepts while preserving coherent temporal relationships. We argue that the cross-attention mechanism in resamplers is promising for addressing these two limitations, since it allows flexible remodeling of visual cues across spatial and temporal dimensions. The key lies in explicitly guiding this remodeling process towards distinct semantics and coherent temporals, which is absent in current encapsulation techniques.

% Solutions
To this end, we propose \textbf{DisCo}, a novel visual encapsulation method that is capable of generating visual tokens with \textbf{dis}tinct semantics and \textbf{co}herent temporal cues, as depicted in~\cref{fig:teaser}(c). DisCo features two principal designs: (\romannumeral1) A Visual Concept Discriminator (VCD) module, which aligns each visual token with a distinct semantic concept. Diverging from previous encapsulation methods that uniformly align all visual tokens with the entire video caption, VCD dynamically assigns different visual tokens to discrete text instances extracted from video descriptions. This approach reduces token redundancy and enhances semantic diversity. (\romannumeral2) A Temporal Focus Calibrator (TFC) module, which aligns the focused instance of each visual token across the temporal dimension. Unlike previous methods that only align visual tokens at video-level, TFC dives into frame-level calibrations between visual tokens and video instances. We introduce a Frame-level Focus Alignment (FFA) loss to guide each visual token to remain aligned with its designated semantic instance throughout each video frame, ensuring temporal coherence across the video. Extensive experiments demonstrate that DisCo achieves state-of-the-art performances on video understanding. Moreover, by reducing information redundancies in visual encapsulation process, DisCo could improve the efficiency of video MLLMs by utilizing 75\% less tokens while maintaining overall performance.

% Summary of Contributions
We summarize our contributions as follows:
\begin{itemize}
    \item We propose DisCo, the first visual encapsulation method that is capable of generating semantically distinct and temporally consistent visual tokens for video LLMs, greatly promoting the quality of visual representations in video-language learning.
    \item In DisCo, a Visual Concept Discriminator (VCD) module is raised to endow visual tokens with unoverlapped semantic concepts, facilitating semantic distinctiveness in visual representations. Additionally, a Temporal Focus Calibrator (TFC) module is introduced to realize frame-level attention on video instances, ensuring the temporal coherence in visual tokens. 
    \item As a plug-and-play design, DisCo is compatible with various video MLLM frameworks. Extensive experiments on multiple baselines demonstrate the superiority and efficiency of DisCo across a wide spectrum of video understanding benchmarks.
\end{itemize}
\section{Related Works}
\label{sec:related works}

\begin{figure*}
    \centering
    \includegraphics[width=1.0\linewidth]{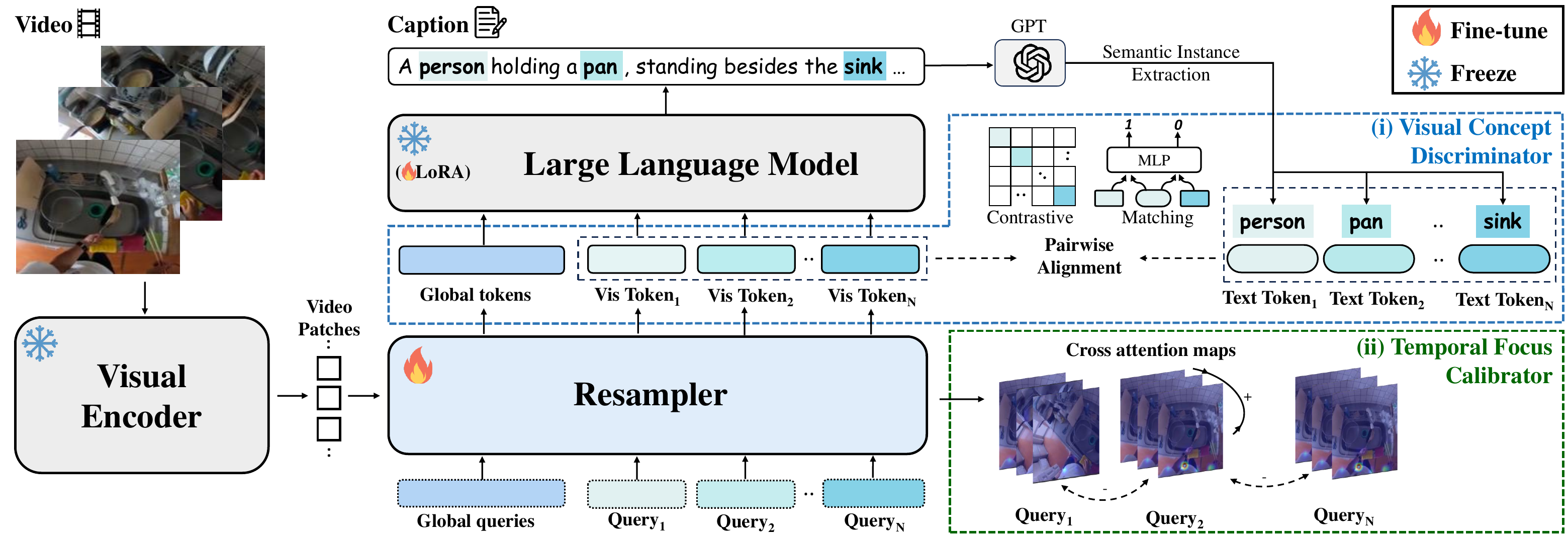}
    \caption{The overall structure of DisCo. DisCo is highlighted by (i) a Visual Concept Discriminator (VCD) module, which conducts a pairwise alignment between visual tokens and semantic concepts, to acquire distinct semantics, and (ii) a Temporal Focus Calibrator (TFC) module, which aligns frame-level focused areas within each visual token, to improve temporal coherence.}
    \label{fig:structure}
\end{figure*}

\noindent\textbf{Multimodal Large Language Models.} \ With the significant advances in Large Language Models (LLMs)~\cite{devlin2018bert,brown2020language,dubey2024llama, longpre2023flan, zheng2023judging}, there is a surge of investigations on exploring Multi-modal Large Language Models (MLLMs)~\cite{alayrac2022flamingo, li2024llava, zhang2024internlm, zhu2023vl}, as they can handle a diverse range of open-ended tasks~\cite{plummer2015flickr30k, hudson2019gqa, lin2014microsoft}. Seminal works like Flamingo~\cite{alayrac2022flamingo} effectively unified the understanding of vision and text modalities, showing impressive performance on a wide range of multi-modal tasks. Recently, a line of open-source MLLMs like LLaVA~\cite{liu2024visual}, Qwen-VL~\cite{wang2024qwen2} and MM-ICL~\cite{zhao2023mmicl} further incorporate visual instruction tuning data~\cite{xu2016msr, li2023mimic, yu2016modeling} to enhance visual dialogue ability. Based on the success of perceiving static images, several studies leverage extensive video-text data corpus~\cite{bain2021frozen, li2023videochat, zhou2018towards} to construct video MLLMs, such as VideoChat~\cite{li2024mvbench}, Video-ChatGPT~\cite{maaz2023video} and InternVideo~\cite{wang2024internvideo2}. Despite their outstanding capabilities in open-world video understanding~\cite{zhao2023antgpt, wu2024star, yu2023anetqa, chen2023egoplan, zheng2024villa}, recent video MLLMs have not yet deeply explored visual connectors, which hold a critical role in deciding the performance and efficiency of MLLMs. In this study, we investigate developing a visual encapsulation method that contributes to a well-performed and efficient video MLLM.

\noindent\textbf{Visual Encapsulation in MLLM.} \ Visual encapsulation is a crucial process in multi-modal large language models that bridges visual encoders with LLMs. Among major visual encapsulation methods, linear projection~\cite{liu2024visual, lin2023video, chen2023minigpt, xu2024pllava, chen2024internvl} is most widely utilized. This design fully preserves visual information, but leads to high computational load due to the large number of visual patches. Another type of encapsulation uses a resampler~\cite{li2023blip, wang2024visionllm, chen2024internvl, ye2024mplug, zhang2023video} to compress the visual patches into a much smaller number of tokens, at the cost of sacrificing the comprehensiveness of visual cues. To achieve token efficiency as well as preserve detailed visual information, works like DeCo~\cite{yao2024deco} and TokenPacker~\cite{li2024tokenpacker} presented token downsampling modules, while Slot-VLM~\cite{xu2024slot} adopted slot attention~\cite{locatello2020object} to capture object-level information. However, the visual representations from these models still lack semantic clarity and temporal coherence. In this work, we address these issues by raising DisCo, a visual encapsulation method that learns semantically distinct and temporally coherent video tokens.
\section{Method}
\label{sec:method}

As illustrated in~\cref{fig:structure}, we propose DisCo, a novel visual encapsulation method designed to generate semantically distinct and temporally coherent visual tokens for video MLLMs. DisCo is highlighted by two primary components: (\romannumeral1) A Visual Concept Discriminator (VCD) module, which aligns a set of visual tokens with a group of semantic concepts in a pairwise manner, achieving distinct semantics. (\romannumeral2) A Temporal Focus Calibrator (TFC) module, which extracts frame-level focused features of visual tokens, and aligns these features across all frames, to ensure coherent temporal attentions. In~\cref{sec:method-video abstractor}, we will first provide the preliminaries on the structure of DisCo. Then, we will introduce the VCD and TFC modules, in \cref{sec:method-vcd} and \cref{sec:method-tfc}, respectively. Finally, we describe the training scheme in~\cref{sec:method-training}.

\subsection{Preliminaries}
\label{sec:method-video abstractor}

In the video MLLM family, a group of models employ resamplers for visual encapsulation. These models are structurally composed of three main components: a visual encoder, a resampler, and a large language model (LLM).

\noindent\textbf{Visual Encoder.} Given a video input sampled into $T$ frames $X=\{x_i\}_{i=1}^T$, a ViT~\cite{dosovitskiy2020image} $\mathcal{V}$ is utilized to extract deep video features $V=\{v_i \in \mathbb{R}^{n \times c}\}_{i=1}^T$. %These features are then concatenated into a feature representation $\hat{V} \in \mathbb{R}^{nT \times c}$.

\noindent\textbf{Resampler.} \ Serving as a bridge between the visual encoder and the LLM, in the resampler (\eg, Q-Former~\cite{li2023blip}), a set of learnable query embeddings $X_q=\{q_i\}_{i=1}^N$ is initialized to interact with video features $V$ through cross-attention layers~\cite{vaswani2017attention}. This interaction produces a set of visual tokens, denoted as $X_v=\{v_i\}_{i=1}^N$, which contain encapsulated visual representations.

\noindent\textbf{Large Language Model (LLM).} \ Large language model acts as a unified platform to process both vision and language inputs, generating natural language answers accordingly. LLM takes the output tokens of resampler $X_v$ as vision input, and a paired text instruction $X_i$ as language input. The entire video LLM is trained by minimizing the negative log-likelihoods of generating the target answer $X_a$:
\begin{equation}
    \mathcal{L}_{llm} = 
        -\mathbb{E}_{X \sim \mathcal{D}}
        \left[\sum_{l=1}^{L} \log p(X_a^l | X_v, X_i^{< l}, X_a^{< l})\right],
    \label{eq:llm loss}
\end{equation}
where $\mathcal{D}$ denotes the training dataset, and $X_i^{< l}$, $X_a^{< l}$ denotes the instruction and answer tokens before the current generated token $X^l$.

\subsection{Visual Concept Discriminator}
\label{sec:method-vcd}

\begin{figure}[!t]
    \centering
    \includegraphics[width=1.0\linewidth]{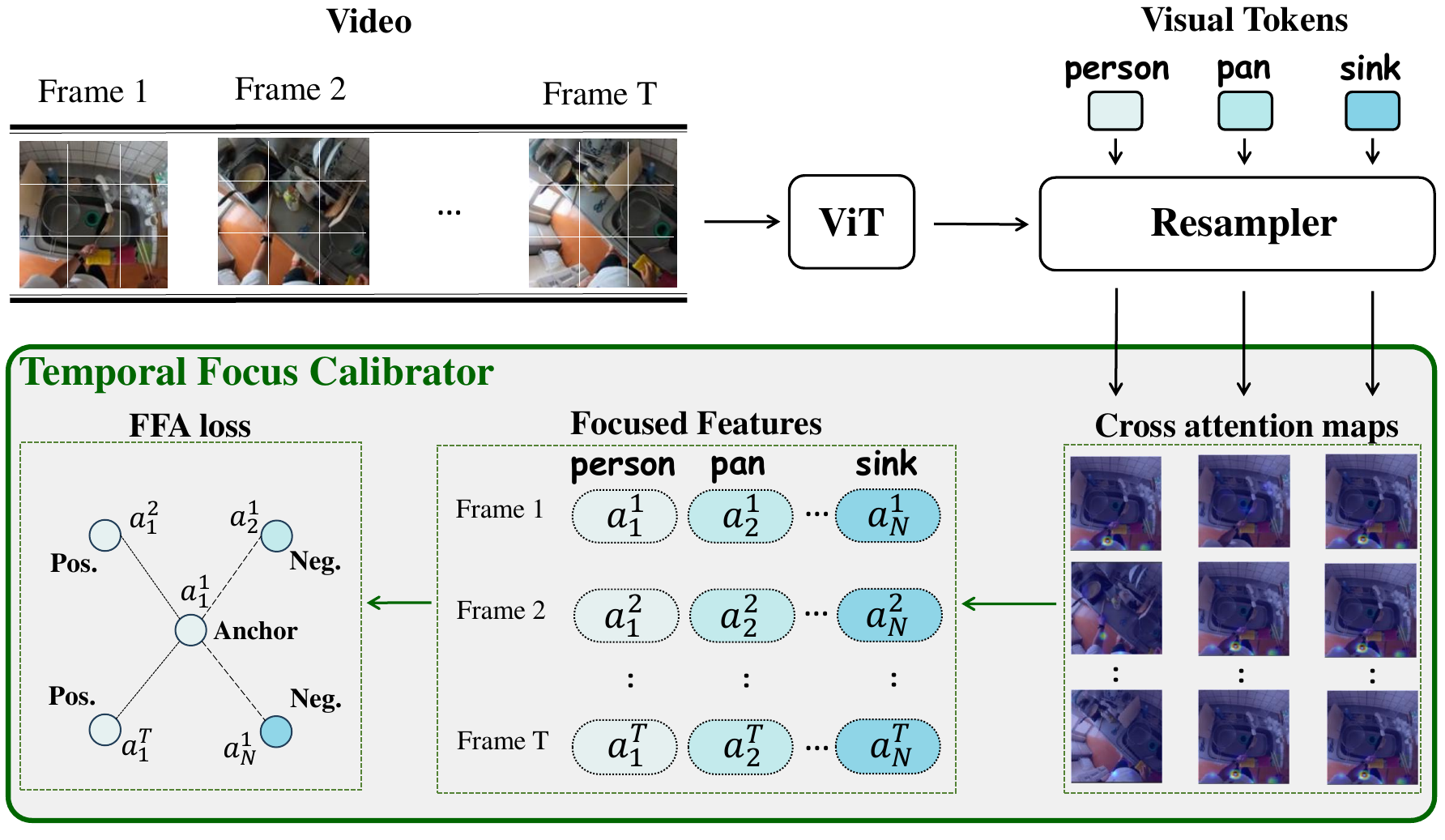}
    \caption{The structure of the TFC module. In TFC, frame-wise focused features are drawn from the cross attention maps in the resampler. Then, the Frame-level Focus Alignment (FFA) loss aligns each frame-wise feature within each visual token, promoting the temporal coherence across video frames. }
    \label{fig:fsfa-structure}
\end{figure}

In existing resampler-based video MLLMs, visual tokens produced by the resampler often endure semantic indistinctness, with multiple tokens representing the same element. We argue that this issue arises from the lack of explicit guidance on the element-wise contents of each visual token, leading to repetitive semantic information among tokens.

To address this problem, we propose a Visual Concept Discriminator (VCD). It distinguishes itself from previous encapsulation methods by explicitly aligning different visual tokens with distinct semantic concepts in a pairwise manner. To implement this pairwise alignment, both visual tokens and semantic concepts are initially divided into multiple groups. As shown in~\cref{fig:structure}, for visual tokens, we reorganize the total of $N$ visual tokens into $N_g$ groups, denoted as $\left\{\hat{v}_i\right\}_{i=1}^{N_g}$, with each group $\hat{v}_i$ comprising $N / N_g$ tokens. For semantic concepts, we leverage GPT-4~\cite{achiam2023gpt} to extract distinct words or phrases that each represent a specific instance in the video caption, forming a set of $M$ semantic concepts. The text embeddings for these semantic concepts are then generated using the resampler's text processing branch, resulting in embeddings $\{\hat{t}_j\}_{j=1}^{M}$.

To achieve a one-to-one alignment between visual token groups and semantic concepts, we perform bipartite matching~\cite{carion2020end} between visual tokens and text embeddings. For simplicity, we assume $N_g=M$. The bipartite matching algorithm determines a permutation of $M$ elements $\hat{\sigma} \in \mathcal{P}_M$, which pairs $i$-th visual token with $\hat{\sigma}(i)$-th semantic concept with lowest cost:
\begin{equation}
    \hat{\sigma} = \mathop{\arg\min}_{\sigma \in \mathcal{P}_M} 
        \sum_{i}^{M} c(v_i, t_{\sigma(i)}),
\end{equation}
where $c(x, y)$ denotes the cosine distance between $x$ and $y$. This assignment is computed using the Hungarian algorithm. In circumstances where $N_g \neq M$, our process yields $\min (N_g, M)$ matching pairs, while leaving the excessive visual or textual elements unused in the VCD module.

\begin{table}[t]
    \centering
    \caption{Comparison of different visual encapsulation methods. DisCo is the first to combine the traits of \textbf{Distinct}: semantic distinction, \textbf{Coherent}: temporal coherence, \textbf{Complete}: information completeness and \textbf{Efficient}: token efficiency.}
    % \vspace{-5pt}
    \setlength\tabcolsep{2mm}
    \begin{adjustbox}{width=\linewidth,center}
    \begin{tabular}{l|cccc}
        \toprule
        \textbf{Encapsulation Methods} & Distinct & Coherent & Complete & Efficient \\
        \midrule
        Linear Projector & \xmark & \xmark & \cmark & \xmark \\
        Resampler & \xmark & \xmark & \xmark & \cmark \\
        \midrule
        DisCo (Ours) & \cmark & \cmark & \cmark & \cmark \\
        \bottomrule
    \end{tabular}
    \end{adjustbox}
    \label{tab:discussion}
% \vspace{-5pt}
\end{table}

Upon establishing the one-to-one matching, we use pairwise losses to facilitate learning the alignment between each pair of visual and semantic features. Matched visual-semantic pairs are treated as positive pairs, while others are considered as negative pairs. Following vision-language alignment techniques in~\cite{radford2021learning, li2022blip}, we apply a visual-semantic pairwise contrastive (VSC) loss and a visual-semantic matching (VSM) loss, denoted as $\mathcal{L}_{vsc}$ and $\mathcal{L}_{vsm}$, respectively. The VSC loss is defined as:
\begin{equation} \small
    \mathcal{L}_{vsc} = -\sum_i^{M} [
        \log \frac{S(v_i, t_{\hat{\sigma}(i)})} 
                {\sum_j S(v_i, t_{\hat{\sigma}(j)})} + 
        \log \frac{S(t_{\hat{\sigma}(i)}, v_i)}
                {\sum_j S(t_{\hat{\sigma}(i)}, v_j)} ],
    \label{eq:vsc loss}
\end{equation}
where $S(v, t)=exp(\frac{v^T t}{\tau |v||t|})$ denotes visual-text similarity score with temperature $\tau$. The VSM loss is expressed as:
\begin{equation}
    \mathcal{L}_{vsm} = \sum_i^{M} 
        \text{CE}(p_{\theta}(v, t), y_{v,t}), (v, t) \sim (v_i, t_j),
    \label{eq:vsm loss}
\end{equation}
where $\text{CE}(p,y)$ denotes the cross-entropy loss between prediction $p$ and ground-truth label $y$. An MLP is utilized to predict $p_{\theta}(v, t)=\mathrm{MLP}([v,t])$.

Since the extracted semantic concepts do not completely contain the original video caption (\eg, the term "holding" is not included as shown in~\cref{fig:structure}), the $N_g$ groups of aligned visual tokens cannot cover complete video information. To ensure comprehensive visual representation, we add a set of global tokens into VCD to capture this uncovered information and preserve the integrity of the visual cues.

\begin{table*}[t]
    \centering
\caption{Performance on video question-answering benchmarks. `val' denotes validation set for PerceptionTest, and `subset' denotes the subset for EgoSchema test set. The best result of each benchmark is boldfaced.
}
\begin{adjustbox}{width=\linewidth,center}
\renewcommand{\arraystretch}{1.1}
\setlength{\tabcolsep}{1.5mm}
% \tablestyle{3pt}{1.0}
\begin{tabular}{lrccccccccc}
\toprule  
\multirow{2}{*}{\textbf{Model}} & \multirow{2}{*}{\textbf{Size}} & \multirow{2}{*}{\textbf{MVBench}} & \multirow{2}{*}{\textbf{STAR}} & {\textbf{PerceptionTest }} & {\textbf{EgoSchema}} & \multirow{2}{*}{\textbf{MLVU}} & \multicolumn{4}{c}{\textbf{VideoMME \textit{(w/o \& w. sub)}}} \\
& & & & val & subset & & overall & short & medium & long \\
% & & &  & & & & & \\
% Average duration (sec) & & 16 & 23 & 180 & 473 & 651 &  1010 \\
% \midrule
% \textit{Proprietary Models} \\
% GPT4-V & - & - & 43.7 & - & - & 59.1  & 49.2 & 59.9 \\
% GPT4-o & - & - & 64.6 & - & 72.2 & 66.7  & 64.6 & 71.9  \\
% Gemini-1.5-Pro & - & - & 60.5 & - & 71.2  & 64.0  & - & 75.0  \\
\midrule
% \textit{Open-Source MLLMs} \\
% InternVL2 & 8B & 66.4 & - & - & -  & - & 54.0 \\
% \color{gray}InternVL2 & \color{gray}76B & \color{gray}69.6 & \color{gray}- & - & \color{gray}-  & \color{gray}- & \color{gray}61.2 \\
Otter-V~\cite{li2023otter} & 7B & 26.8 & - & - & - & 16.7 & - & - & - & - \\
VideoLLaMA~\cite{zhang2023video} & 7B & 33.6 & 26.3 & 36.5 & 25.6 & - & 26.5/37.1 & 25.7/27.8 & 25.1/35.6 & 28.6/38.1 \\
VideoChat2~\cite{li2024mvbench} & 7B & 35.5 & 59.0 & - & 64.6 & - & 39.5/43.8 & 48.3/52.8 & 37.0/39.4 & 33.2/39.2 \\
LLaMA-VID~\cite{li2025llama} & 7B & 41.3 & - & - & - & 18.1 & 25.9/ - & - & - & - \\
VideoLLaVA~\cite{lin2023video} & 7B & 43.0 & - & - & - & 29.3 & 39.9/41.6 & 45.3/46.1 & 38.0/40.7 & 36.2/38.1 \\
LLaVA-Mini~\cite{zhang2025llava} & 7B & 44.5 & - & - & - & 42.8 & - & - & - & - \\
LongLLaVA~\cite{wang2024longllava} & 9B & 49.1 & - & - & - & - & 43.7/ - & - & - & - \\
ShareGPT4Video~\cite{chen2024sharegpt4video} & 8B & 51.2 & - & - & - & 34.2 & 39.9/43.6 & 48.3/53.6 & 36.3/39.3 & 35.0/37.9 \\
LLaVA-NeXT-Video~\cite{zhang2024llava} & 7B & 53.1 & 35.5 & 48.8 & 49.1 & - & 37.3/43.7 & 39.3/47.8 & 38.9/46.9 & 33.9/36.2 \\
VideoLLaMA2~\cite{cheng2024videollama} & 7B & 54.6 & 57.2 & 51.4 & 51.7 & 48.5 & \textbf{47.9}/50.3 & 54.3/56.1 & 44.3/47.4 & 40.1/45.7 \\
% \color{gray}VideoLLaMA2 & \color{gray} 72B & \color{gray}62.0 & \color{gray}57.5 & \color{gray}63.9 & \color{gray}-  &\color{gray}- & \color{gray}62.4/64.7 \\
% mPLUG-Owl3 & 7B & 54.5 & - & - & - & - & 53.5 & & & \\
% QwenVL2 & 7B & 67.0 & 62.3  & 66.7 & -  & - & 63.3 \\
% \color{gray}QwenVL2 & \color{gray}72B & \color{gray}73.6 & \color{gray}68.0  & \color{gray}77.9 & \color{gray}-  & \color{gray}- & \color{gray}71.2 \\
VideoChat2-HD~\cite{li2024mvbench} & 7B & 62.3 & 63.9 & 54.3 & 65.6 & 47.9 & 45.3/55.7 & 53.4/59.2 & 47.3/54.0 & 37.1/46.7 \\
% LLaVA-OneVision & 7B & 56.7 & 57.1  & 60.1 & 56.3  & 64.7 & 58.2 \\
% \color{gray} LLaVA-OneVision & \color{gray} 72B &\color{gray}59.4 & \color{gray}66.9  & - & \color{gray}61.3  & \color{gray}68.0 & \color{gray}66.2\\
% VideoChat-TPO & 7B & 66.8 & - & - & -  & 54.7 & - \\
\midrule
% \textit{Open-Source Long Video MLLMs} \\

% LLaMA-VID & 7B & 2 & 41.9 & 44.6 & - & - & 33.2 & 25.9 \\
% LongVILA & 7B & 196 & - & - & 67.7 & -  & - & 57.5 \\
% LongVA & 7B & 144 & - & - & - & -  & 56.3 & 52.6 \\
% LongLLaVA & 9B & 144 & 49.1 & - & - & -  & -& 43.7 \\
% LongVU & 7B & 64 & 66.9 & - & 67.6 & -  & 65.4 & - \\
% VideoChat-Flash & 7B & 16 & 73.2 & 75.6 & - & 64.2  & 74.5 & 64.0 \\
ST-LLM~\cite{liu2025st} & 7B & 54.7 & 56.7 & 49.5 & 55.2 & 46.7 & 40.6/- & 49.9/- & 40.2/- & 31.5/- \\
\rowcolor{mygray}\textbf{ST-LLM+DisCo} & 7B & 58.0 & 60.1 & 54.4 & 59.8 & 48.6 & 42.1/- & 51.8/- & 39.6/- & 34.8/- \\
InternVideo2~\cite{wang2024internvideo2} & 7B & 60.3 & 64.5 & 52.6 & 64.4 & 43.9 & 41.7/51.7 & 50.3/56.7 & 37.4/50.1 & 37.3/48.0 \\
\rowcolor{mygray}\textbf{InternVideo2+DisCo} & 7B & 63.3 & 72.7 & 61.7 & 66.2 & 46.7 & 42.9/52.8 & 53.0/59.5 & 38.7/50.0 & 37.0/48.7 \\
InternVideo2-HD~\cite{wang2024internvideo2} & 7B & 66.3 & 75.7 & 62.4 & 67.0 & 47.1 & 46.3/56.7 & 54.5/59.5 & 42.4/55.3 & 42.0/55.3 \\
\rowcolor{mygray}\textbf{InternVideo2-HD+DisCo} & 7B & \textbf{68.2} & \textbf{77.7} & \textbf{67.4} & \textbf{72.2} & \textbf{49.5} & 47.4/\textbf{57.9} & \textbf{55.8}/\textbf{61.3} & \textbf{43.8}/\textbf{56.1} & \textbf{42.7}/\textbf{56.2} \\

\bottomrule
\end{tabular}
\end{adjustbox}
\label{tab:main_result}
\end{table*}

\subsection{Temporal Focus Calibrator}
\label{sec:method-tfc}
\vspace{-0.3mm}

Despite the improvements brought by VCD, existing resamplers still face the challenge of temporal incoherence. Delving into their mechanisms, it is revealed that each visual token is uniformly attended to video patches from all frames. This approach fails to ensure that each visual token consistently focuses on every individual video frame. As a solution, we introduce a Temporal Focus Calibrator (TFC) module, which pioneers frame-level calibration for resamplers in video MLLMs. The primary aim of TFC is to explicitly supervise each visual token to focus on its corresponding semantic concepts in each frame.

As illustrated in~\cref{fig:fsfa-structure}, the initial step of TFC involves extracting the focused features from the cross-attention maps between visual tokens and the video features extracted by ViT. Specifically, for the $i$-th token, we denote its cross-attention maps with the $t$-th video frame as $\{C_k^t \in \mathbb{R}^{h \times w}\}_{k=1}^{L_c}$, where $L_c$ is the total number of cross-attention layers. Then, the attention features are as follows:
\begin{small}
\begin{equation}
    a_i^t = \text{AvgPool}(\frac{1}{L_c} \sum_{k=1}^{L_c} C_k^t \cdot V^t), i=1 \sim N_g, t=1 \sim T,
    \label{eq:attention feature}
\end{equation}
\end{small}
\vspace{-0.3mm}where $\text{AvgPool}(\cdot)$ denotes average pooling along the spatial dimensions, and $V^t$ is the video feature of $t$-th frame.

To achieve alignment of frame-wise attention features within each visual token, we present a Frame-level Focus Alignment (FFA) loss: given attention feature $a_i^t$ as an anchor, FFA loss pulls $a_i^t$ closer to the attention features of other frames within the $i$-th token, while pushes $a_i^t$ apart from the attention features of other tokens. Moreover, to improve the stability of frame-wise attention features (particularly in cases where an object may temporarily disappear in some frames), we utilize the feature centroid of each query, defined as $\overline{a}_i=\frac{1}{T} \sum_{t=1}^T a_i^t$. The centroid feature provides a more robust reference for alignment in the FFA loss. Finally, the loss is formulated as follows:
\begin{small}
\begin{equation}
    \mathcal{L}_{ffa} = - \sum_i^{N_g} \sum_t^{T}
        \left[
        \log \frac{S(\overline{a}_i, a_i^t)}{\sum_j S(\overline{a}_i, a_j^t)} + 
        \log \frac{S(a_i^t, \overline{a}_i)}{\sum_j S(a_i^t, \overline{a}_j)}
        \right].
    \label{eq:fsc loss}
\end{equation}
\end{small}

\subsection{Training}
\label{sec:method-training}
Following standard training strategies of MLLMs, our training process consists of two stages. Stage 1 focuses on vision-text alignment. In this stage, we leverage a substantial dataset of visual dense captions to align the visual tokens of DisCo with the LLM. Additionally, the VCD and TFC modules are incorporated in this stage. The total training loss is formulated as:
\begin{small}
\begin{equation}
    \mathcal{L}_{stage1} = 
        \mathcal{L}_{llm} + 
        \lambda_{vsc} \mathcal{L}_{vsc} +
        \lambda_{vsm} \mathcal{L}_{vsm} +
        \lambda_{ffa} \mathcal{L}_{ffa}.
    \label{eq:stage1 loss}
\end{equation}
\end{small}where $\lambda_{vsc}$, $\lambda_{vsm}$, and $\lambda_{ffa}$ are weight parameters. After completing Stage 1, we advance to Stage 2, the instruction tuning stage. In this stage, we utilize a diverse set of image and video caption and question-answer (QA) data to equip the model with strong instruction following ability.

\subsection{Discussion}
\label{sec:method-discussion}

Now we illustrate the difference between DisCo and existing visual encapsulation methods. As shown in~\cref{tab:discussion}, all previous methods endure indistinctness in token semantics, and incoherence in temporal modeling. Instead, DisCo encapsulates the visual token with two defining attributes: (\romannumeral1) \textbf{Semantic distinction:} each visual token represents unoverlapped instances, possessing clear semantic difference. (\romannumeral2) \textbf{Temporal coherence:} each visual token attends the dynamics of its corresponding instance at every frame. Moreover, by reducing overlapped semantics, DisCo achieves: (\romannumeral3) better \textbf{Information completeness} by covering more visual elements, and (\romannumeral4) \textbf{Token efficiency} by utilizing less tokens to represent the same amount of visual cues.
\section{Experiments}
\label{sec:experiments}

\begin{table}[!t]
\centering
\caption{Comparison with state-of-the-art methods on video conversation benchmarks. `CI', `DO', `CU', `TU', and `CO' denote `Correctness of Information', `Detail Orientation', `Context Understanding', `Temporal Understanding', and `Consistency'.}
\begin{adjustbox}{width=\linewidth,center}
\renewcommand{\arraystretch}{1.1}
\tablestyle{5pt}{1.0}
\begin{tabular}{l|ccccc|c}
\toprule
\textbf{Model} & CI & DO & CU & TU & CO & Avg \\
\midrule
VideoLLaMA~\cite{zhang2023video} & 1.96 & 2.18 & 2.16 & 1.82 & 1.79 & 1.98 \\
VideoChatGPT~\cite{maaz2023video} & 2.40 & 2.52 & 2.62 & 1.98 & 2.37 & 2.38 \\
VideoChat2~\cite{li2024mvbench} & 3.02 & 2.88 & 3.51 & 2.66 & 2.81 & 2.88 \\
LLaMA-VID~\cite{li2025llama} & 2.96 & 3.00 & 3.53 & 2.46 & 2.51 & 2.89 \\
LLaVA-Mini~\cite{zhang2025llava} & 2.97 & 2.99 & 3.61 & 2.48 & 2.67 & 2.94 \\
Chat-UniVi~\cite{jin2024chat} & 2.89 & 2.91 & 3.46 & 2.89 & 2.81 & 2.99 \\
\midrule
% ST-LLM & & & & & &  \\
% \rowcolor{mygray}
% \textbf{ST-LLM+DisCo} & & & & & & \\
InternVideo2~\cite{wang2024internvideo2} & 2.88 & 2.53 & 3.20 & 2.51 & 2.67 & 2.76 \\
\rowcolor{mygray}
\textbf{InternVideo2+DisCo} & 3.13 & 2.65 & 3.42 & 2.56 & 2.89 & 2.93 \\
InternVideo2-HD~\cite{wang2024internvideo2} & 3.14 & 2.74 & 3.53 & 2.52 & 2.85 & 2.96 \\
\rowcolor{mygray}
\textbf{InternVideo2-HD+DisCo} & \textbf{3.36} & \textbf{3.20} & \textbf{3.76} & \textbf{2.80} & \textbf{3.10} & \textbf{3.24} \\
\bottomrule
\end{tabular}
\end{adjustbox}
\label{tab:videochatgpt}
\end{table}

\begin{figure*}
    \centering
    \includegraphics[width=1.0\linewidth]{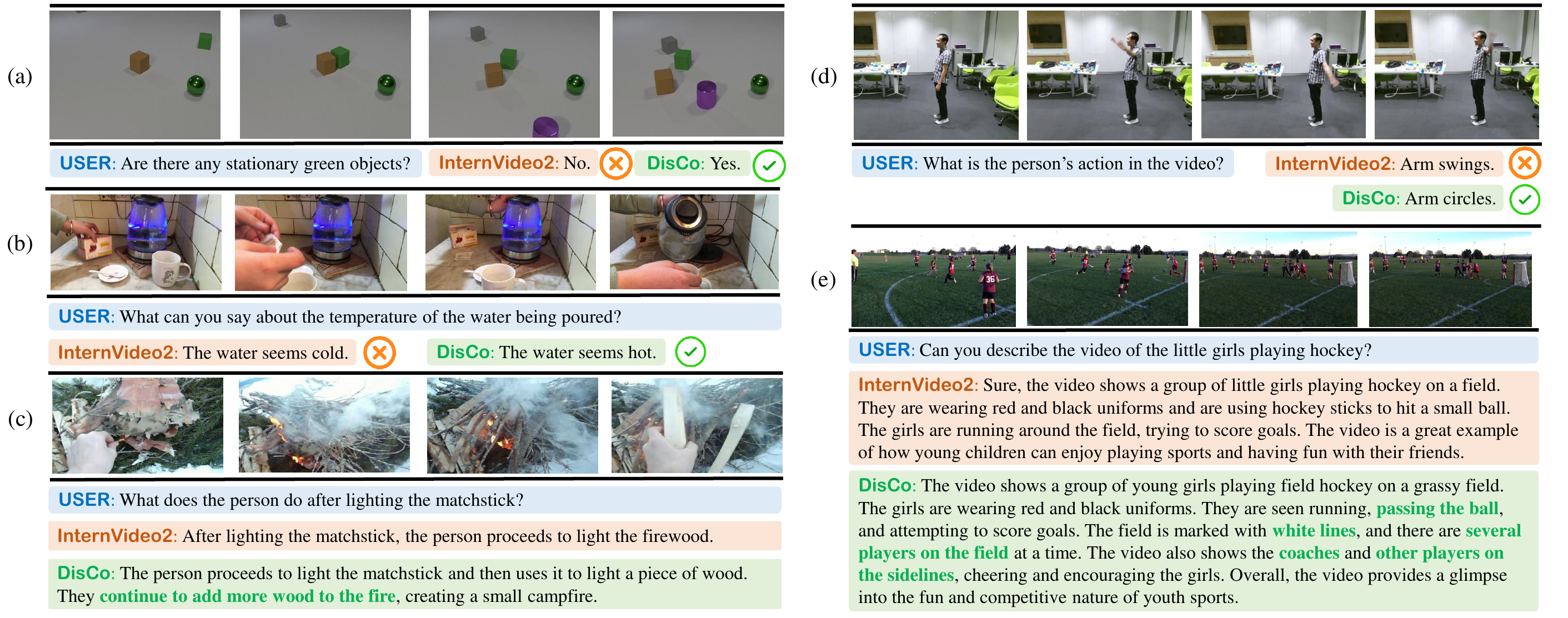}
    \caption{Qualitative examples of video understanding. Utilizing DisCo, video MLLMs achieve (a)(b) better correctness, (c)(d) stronger temporal coherence and (e) richer details in video captioning and QA tasks.}
    \label{fig:qa-example}
\end{figure*}

\paragraph{Implementation Details.} DisCo functions as a plug-and-play module, designed to generally enhance resampler-based video MLLMs. To assess its integration capabilities across different frameworks, we implemented DisCo on two video MLLMs: ST-LLM~\cite{liu2025st} and InternVideo2~\cite{wang2024internvideo2}. ST-LLM employs the ViT-G/14 model from EVA-CLIP~\cite{fang2023eva} as its visual encoder and utilizes Vicuna-7B-v1.1~\cite{chiang2023vicuna} as its LLM. InternVideo2 utilizes InternVideo2-1B as its visual encoder and Mistral-7B~\cite{jiang2023mistral} for LLM. Both models incorporate Q-Former~\cite{li2022blip} as the visual connector. Throughout both training stages, we freeze the visual encoder, update the resampler, and fine-tune the LLM using LoRA~\cite{hu2021lora}.

In implementing DisCo, for ST-LLM, $8$ of the $32$ pretrained query tokens in the resampler are designated as global tokens. The remaining $24$ tokens are distributed across $N_g=12$ visual token groups, each comprising $2$ tokens, to ensure comprehensive coverage of each semantic concept. For InternVideo2, $32$ of the $96$ tokens are assigned as global tokens, and the rest are set into $N_g=16$ groups. In~\cref{eq:stage1 loss}, we set $\lambda_{vsc}$, $\lambda_{vsm}$ and $\lambda_{ffa}$ at $1.0$.

% \subsection{Datasets and Evaluation Metrics}

\noindent\textbf{Datasets.} We adopt a wide scope of video captioning and question-answering (QA) data sources for the training of DisCo. In stage 1, we utilize 900K video dense captions from ShareGPTVideo~\cite{zhang2024direct}, as well as 23K image captions from LLaVA~\cite{liu2024visual}. In stage 2,  our approach aligns with the instructional tuning protocols inherent to the foundational Video MLLMs upon which DisCo is based. Specifically, for the ST-LLM-based DisCo, we incorporate WebVid~\cite{bain2021frozen}, NexT-QA~\cite{xiao2021next}, CLEVRER~\cite{yi2019clevrer}, Kinetics-710~\cite{kay2017kinetics} and Something-Something-v2~\cite{goyal2017something}. For the InternVideo2-based DisCo, we adhere to the recipe used in VideoChat2~\cite{li2024mvbench}.

\noindent\textbf{Evaluation Metrics.} \ For video question-answering (QA) benchmarks, the accuracy of the model's responses is assessed using multiple-choice formats. This close-ended approach enhances objectivity and fairness for evaluation. For video conversation benchmarks, we utilize GPT~\cite{achiam2023gpt} to assign scores for each answer, enabling multi-angled assessments such as detailedness and consistency.

\subsection{Comparison with State-of-the-arts}

We present quantitative evaluations of our proposed DisCo in comparison to state-of-the-art methods across a broad array of video QA benchmarks, including: (1) Short video benchmarks STAR~\cite{wu2024star} and PerceptionTest~\cite{patraucean2024perception} with focus on fine-grained visual details. (2) Long video benchmarks EgoSchema~\cite{mangalam2023egoschema} and MLVU~\cite{zhou2024mlvu}, stressing complex temporal relationships. (3) Comprehensive benchmarks MVBench~\cite{li2024mvbench} and VideoMME~\cite{fu2024video}, covering diverse video QA tasks. To test the capability of DisCo on video conversations, we also validate DisCo on the VideoChatGPT-Bench~\cite{maaz2023video}. 

\begin{table}[]
\vspace{-4mm}
    \centering
    \caption{Ablations on the key components of DisCo. `SFT' denotes using the same training corpus as DisCo to directly fine-tune the baseline model. `VCD (w)' and `TFC (w)' denotes adding VCD and TFC to the baseline, respectively. EgoSchema is validated on the \textit{subset}.}
    \resizebox{1.0\linewidth}{!}{
    \tablestyle{13pt}{1.0}
    \begin{tabular}{l|ccc}
    \toprule
    Methods & MVBench & STAR & Egoschema \\
    \midrule
    Baseline & 66.3 & 75.7 & 67.0 \\
    SFT & 66.5 & 76.3 & 68.2 \\
    VCD (w) & 67.6 & 77.5 & 71.8 \\
    TFC (w) & 67.1 & 77.4 & 70.4 \\
    \midrule
    \rowcolor{mygray}
    \textbf{DisCo} & \textbf{68.2} & \textbf{77.7} & \textbf{72.2} \\
    \bottomrule
    \end{tabular}
    }
    \label{tab:ablation-component}
\end{table}

\begin{figure}[!t]
    \centering
    \includegraphics[width=0.8\linewidth]{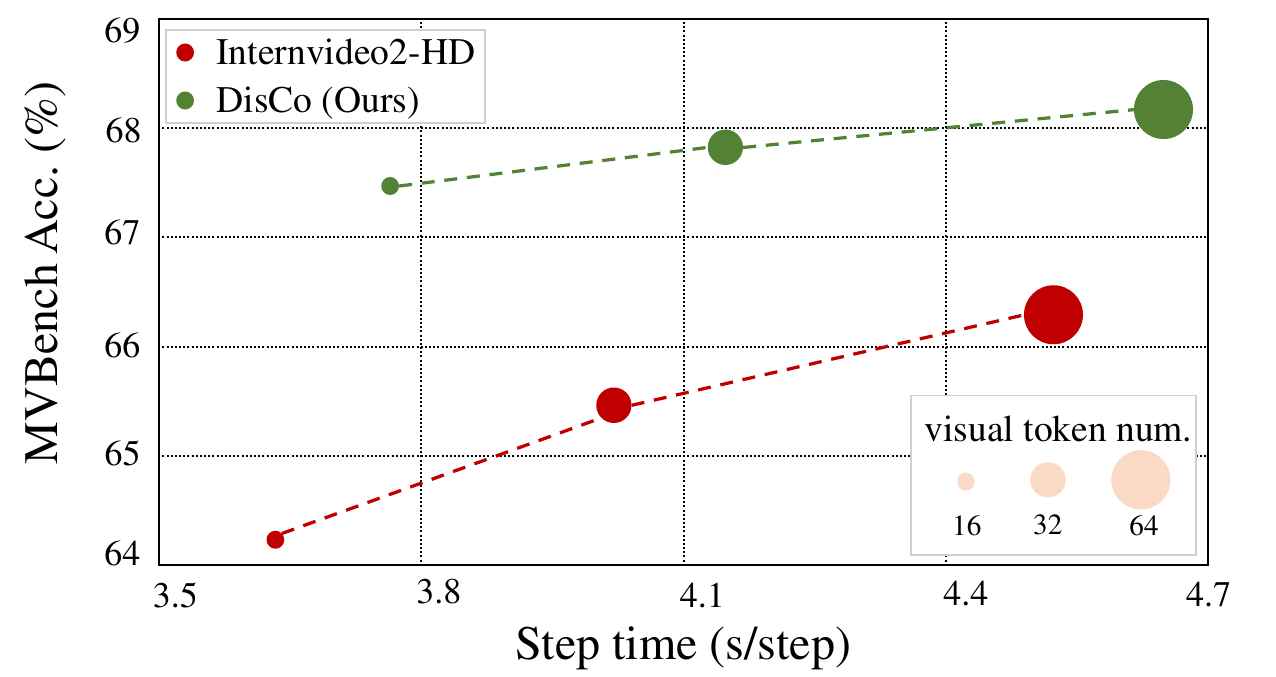}
    \caption{Performance and efficiency under different number of visual tokens. We report performance on MVBench. Efficiency is measured by the average time of each training step.}
    \label{fig: token efficiency}
\vspace{-3mm}
\end{figure}

As depicted in~\cref{tab:main_result}, DisCo consistently enhances the performance of video MLLMs on the video QA benchmarks with various video lengths, question granularity and task diversity. Moreover, the introduction of DisCo consistently improves the performance of ST-LLM and InternVideo2. As shown in~\cref{tab:videochatgpt}, DisCo consistently outperforms current state-of-the-art methods on video conversation benchmarks. This result validates the comprehensive enhancement DisCo brings to video MLLMs.

In~\cref{fig:qa-example}, we present qualitative examples of DisCo. (a) and (b) shows that DisCo possesses stronger abilities on grabbing detailed visual cues like object colors and water steam, leading to better results on detailed understanding. (c) and (d) proves that DisCo captures temporal events more coherently (add wood to fire, arm movements), performing better on temporal reasoning. (e) shows that DisCo yields video captions with more sufficient and fine-grained visual details, demonstrating its superiority on video captioning.

\begin{figure*}[t]
    \centering
    \includegraphics[width=0.9\linewidth]{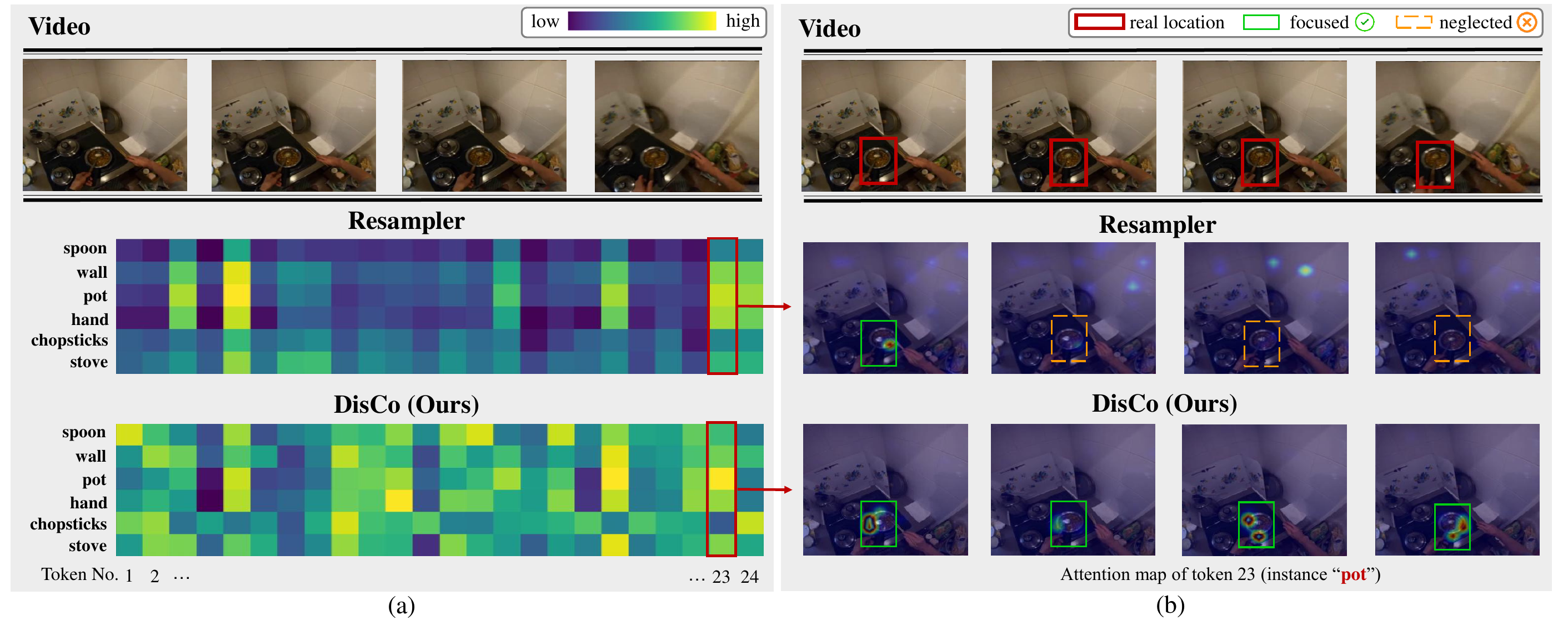}
    \caption{(a) Similarity matrix between visual tokens and text instances. Resamplers exhibit severe semantic redundancy across tokens, while DisCo achieves distinct semantics. (b) Attention maps between the visual token representing `pot' and each video frame. Resamplers fail to consistently attend to the instance `pot', while DisCo attends to it in every frame, demonstrating good temporal coherence.}
    \label{fig:visualization}
    \vspace{-5pt}
\end{figure*}

\subsection{Ablation Studies}

We conduct a thorough analysis on the effectiveness of the primary components and key designs in DisCo. More ablations could be found in the supplementary materials.

\begin{table}[]
    \centering
    \caption{Ablations on the VCD module. `$N_g$' denotes the number of visual token groups, and `$N/N_g$' denotes the number of tokens in each group. `Global' stands for global tokens.}
    \resizebox{1.0\linewidth}{!}{
    \tablestyle{7pt}{1.0}
    \begin{tabular}{c|c|c|cc|cc}
    \toprule
    $N_g$ & $N/N_g$ & Global & $\mathcal{L}_{vtc}$ & $\mathcal{L}_{vtm}$ & MVBench & STAR \\
    \midrule
    $4$ & $16$ & \Checkmark & \Checkmark & \Checkmark & 66.5 & 77.1 \\
    $64$ & $1$ & \Checkmark & \Checkmark & \Checkmark & 67.3 & 77.5 \\
    $16$ & $4$ & & \Checkmark & \Checkmark & 65.8 & 76.4 \\
    \midrule
    $16$ & $4$ & \Checkmark & \Checkmark & & 67.4 & 76.7 \\
    $16$ & $4$ & \Checkmark & & \Checkmark & 66.8 & 75.7 \\
    \midrule
    \rowcolor{mygray}
    $\boldsymbol{16}$ & $\boldsymbol{4}$ & \Checkmark & \Checkmark & \Checkmark & \textbf{68.2} & \textbf{77.7} \\
    \hline
    \end{tabular}
    }
    \label{tab:ablation-vspa}
\end{table}

\begin{table}[]
    \centering
    \caption{Ablations on the TFC module. `Frame-wise feat' denotes only using frame-wise attention features when implementing the FFA loss (\cref{eq:fsc loss}). `Feat. centroid' denotes using frame-level average in the FFA loss. EgoSchema is validated on the \textit{subset}.}
    \resizebox{1.0\linewidth}{!}{
    \tablestyle{9pt}{1.0}
    \begin{tabular}{l|ccc}
    \toprule
    Methods & MVBench & STAR & EgoSchema \\
    \midrule
    Frame-wise feat. & 67.7 & 76.4 & 71.0 \\
    \rowcolor{mygray}
    \textbf{Feat. centroid (DisCo)} & \textbf{68.2} & \textbf{77.7} & \textbf{72.2} \\
    \bottomrule
    \end{tabular}
    }
    \label{tab:ablation-fsfa}
\end{table}

\noindent\textbf{Effectiveness of major components.} \ The implementation of DisCo comprises two key components: the Visual Concept Discriminator (VCD) and the Temporal Focus Calibrator (TFC) modules. To assess the effectiveness of each component, we conduct an ablation on these modules. The results, as shown in~\cref{tab:ablation-component}, indicate that both the VCD and TFC modules contribute significant performance improvements across all three benchmarks. 

Furthermore, to ensure that these performance gains originate from the module designs rather than the integration of new data, we utilize the same training corpus as DisCo to directly fine-tune the baseline model, resulting in the SFT model. From the results presented in~\cref{tab:ablation-component}, it is evident that both VCD and TFC achieve higher accuracy compared to SFT by a substantial margin, thereby strongly affirming the efficacy of our component designs.

\noindent\textbf{Improvement on Token Efficiency.} \ As DisCo contributes to alleviating information redundancy in the visual tokens, we explore the potentials of DisCo on improving token efficiency. To this end, we conduct experiments on InternVideo2-HD by varying the number of local visual tokens. From the results in~\cref{fig: token efficiency}, we can conclude that DisCo could maintain its performance when the token number decreases, and a 16-token DisCo even outperforms a traditional resampler with 64 tokens. This proves that DisCo holds great promise on mitigating training and inference costs. Meanwhile, DisCo only introduces minor training consumptions over resamplers, with training time increasing by less than $5\%$ when token numbers are the same.

\noindent\textbf{Ablations on key designs of VCD.} \ The VCD module is designed to mitigate semantic redundancy in visual tokens by aligning group-wise visual tokens with diverse text instances. Our investigations reveal that the number of groups ($N_g$) and the number of tokens per group significantly impact VCD's performance.  As demonstrated in~\cref{tab:ablation-vspa}, reducing the number of groups to $N_g=4$ results in a performance decrease of $1.7\%$ on MVBench, indicating that a limited number of discrete visual token groups impairs the model's ability to capture rich semantic details. Conversely, increasing $N_g$ to $64$ also leads to a performance decline, possibly due to insufficient tokens per group, which compromises the informational completeness of visual tokens for each semantic concept. To this end, we choose the optimal $N_g$ as $16$. This setting guarantees that the visual token groups could cover most of the text samples (with an average instance number of $9.96$, while only $4.95\%$ excess $16$ instances), while also making sure there are not too many unused token groups during training. In addition, VCD incorporates a set of global tokens aimed at capturing global information that may be overlooked by the visual token groups. \cref{tab:ablation-vspa} shows that the existence of global tokens is crucial for DisCo to achieve higher performance.

To further illustrate the effectiveness of VCD, in~\cref{fig:visualization}(a), we present the similarity matrix between visual tokens and semantic concepts. It is clear from the visualization that the tokens from resamplers exhibit severe redundancy, with multiple tokens aligning to `wall', `pot' and `hand', while instances like `spoon' and `chopsticks' are almost ignored by all tokens. In contrast, DisCo guides different visual tokens to highlight distinct semantics, and endows visual tokens with more comprehensive representation of the video content.

In~\cref{tab:ablation-vspa}, we also validate the necessity of introducing VSC and VSM losses in VCD training. Performance on MVBench declines by $1.4\%$ and $0.8\%$ when only using VSC or VSM loss, respectively. This proves the effectiveness of utilizing both losses.

\noindent\textbf{Ablations on key designs of TFC.} \ In the TFC module, to provide a robust foundation for calculating the FFA loss, we employ feature centroids derived from each visual token for contrastive learning. To assess the effectiveness of this approach, we compare the performance of using feature centroids across frames against using frame-wise features. As presented in~\cref{tab:ablation-fsfa}, the use of feature centroids consistently yields performance improvements across all three evaluation benchmarks. These results underscore the effectiveness of employing feature centroids to enhance temporal consistency in visual token alignment.

To better demonstrate the effectiveness of TFC, we visualize the cross-attention maps between a visual token group and all video frames in~\cref{fig:visualization}(b). We can observe that when using resamplers, the token that highlights the instance `pot' only attends to the pot in the first two frames. In the remaining frames, the pot is neglected by the resampler's attention. Conversely, DisCo consistently tracks the pot across all frames. This illustrates the effectiveness of utilizing TFC on improving the temporal coherence of visual tokens.
\section{Conclusion}
\label{sec:conclusion}
This paper proposes DisCo, a visual encapsulation method that first builds \textit{semantically distinct} and \textit{temporally coherent} visual tokens for video MLLMs. By incorporating a novel Visual Concept Discriminator (VCD) module and a Temporal Focus Calibrator (TFC) module, DisCo generates visual tokens with distinct semantic information and robust temporal coherence. Extensive experiments verify that DisCo attains state-of-the-art performance and remarkable efficiency on diverse video understanding benchmarks.

% \noindent\textbf{Acknowledgements.}

\clearpage
% \setcounter{page}{1}
% \renewcommand{\thefigure}{S\arabic{figure}}
% \setcounter{figure}{0}
% \renewcommand{\thetable}{S\arabic{table}}
% \setcounter{table}{0}
% \renewcommand\thesection{\Alph{section}}
% \setcounter{equation}{0}
% \renewcommand\theequation{A.\arabic{equation}}
% \maketitlesupplementary
\appendix

\section{Details of Training}

In~\cref{tab:param-stllm} and~\cref{tab:param-internvideo}, we list the hyper-parameters we adopt for the training of DisCo. In \textbf{\textit{Stage 1}}, for the ST-LLM~\cite{liu2025st} basd DisCo, since ST-LLM did not adopt a pre-training stage, we set the stage 1 hyper-parameters according to their instruction tuning stage. Specifically, following common MLLM pre-training approaches, we adopt larger batch size and larger learning rates. For InternVideo2~\cite{wang2024internvideo2} based DisCo, we follow the hyper-parameter setting of their video-text pretraining stage. In \textbf{\textit{Stage 2}}, we use diverse video conversation data for instruction tuning. For this stage, we follow the hyper-parameter settings of the instruction tuning stage in ST-LLM and InternVideo2, accordingly.

\begin{table}[ht]
    \centering
    \caption{Hyperparameter settings for the training of DisCo based on ST-LLM~\cite{liu2025st} framework.}
    \resizebox{1.0\linewidth}{!}{
    \tablestyle{11pt}{1.0}
    \begin{tabular}{l|cc}
    \toprule
    \multicolumn{3}{c}{\textbf{\textit{ST-LLM}}} \\
    \midrule
    Hyperparameters & Stage 1 & Stage 2 \\
    \midrule
    input frame & $8$ & $8$ \\
    input resolution & $224$ & $224$ \\
    batch size & $512$ & $128$ \\
    total epochs & $1$ & $2$ \\
    % warmup epochs & $1$ & $1$ \\
    learning rate & $1e\mbox{-}4$ & $2e\mbox{-}5$ \\
    learning rate schedule & \multicolumn{2}{c}{cosine decay} \\
    \bottomrule
    \end{tabular}
    }
    \label{tab:param-stllm}
\end{table}

\begin{table}[ht]
    \centering
    \caption{Hyperparameter settings for the training of DisCo based on InternVideo2~\cite{wang2024internvideo2} framework.}
    \resizebox{1.0\linewidth}{!}{
    \tablestyle{11pt}{1.0}
    \begin{tabular}{l|cc}
    \toprule
    \multicolumn{3}{c}{\textbf{\textit{InternVideo2}}} \\
    \midrule
    Hyperparameters & Stage 1 & Stage 2 \\
    \midrule
    input frame & $8$ & $8$ \\
    input resolution & $224$ & $224$ \\
    batch size & $1024$ & $256$ \\
    total epochs & $1$ & $1$ \\
    % warmup epochs & $$ & $$ \\
    learning rate & $1e\mbox{-}4$ & $2e\mbox{-}5$ \\
    learning rate schedule & \multicolumn{2}{c}{cosine decay} \\
    \bottomrule
    \end{tabular}
    }
    \label{tab:param-internvideo}
\end{table}

\begin{figure}[]
    \centering
    \includegraphics[width=1.0\linewidth]{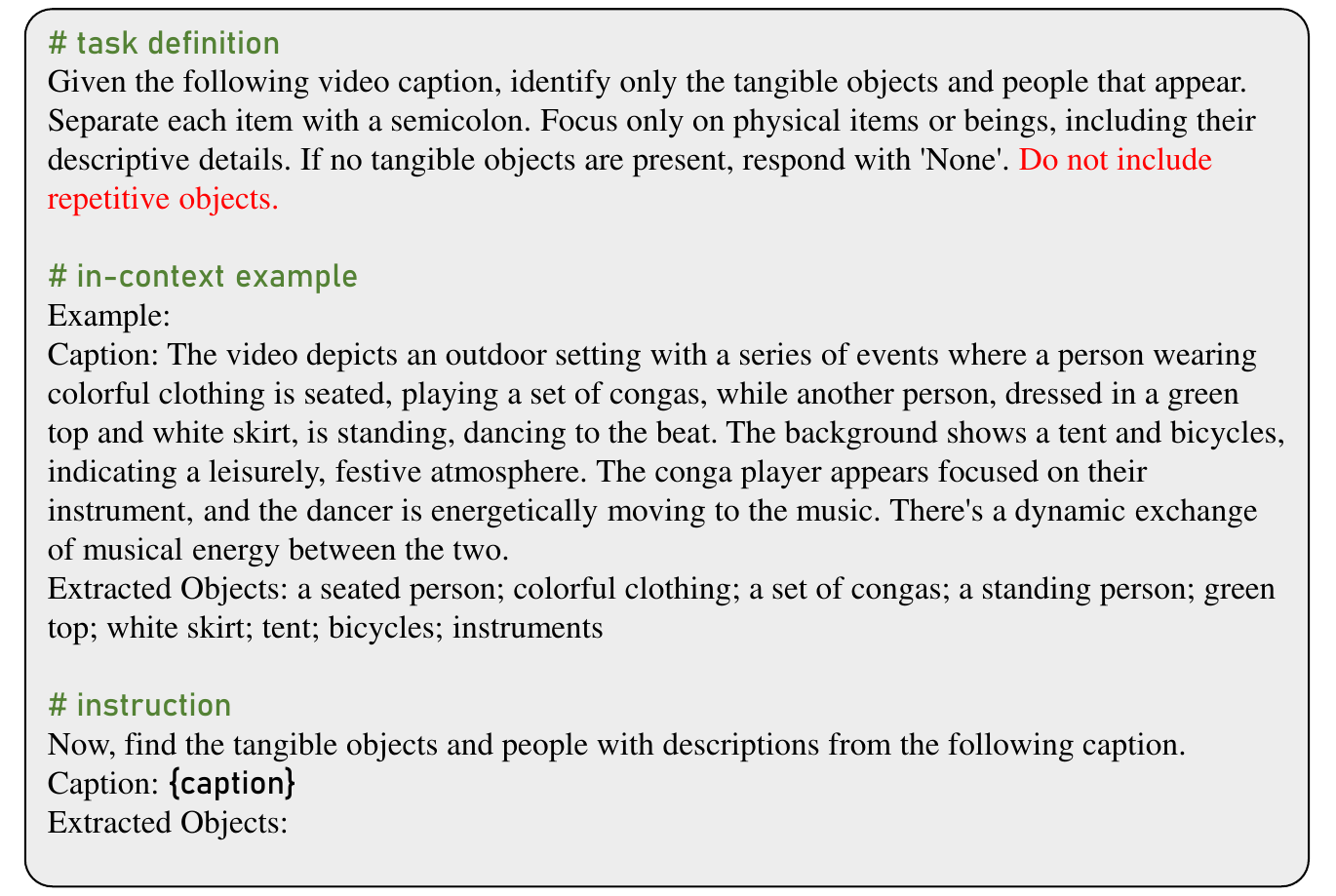}
    \caption{The prompt we used to guide GPT-4 to perform the semantic instance extraction task.}
    \label{fig:supp-prompt}
\end{figure}

\section{Details of Semantic Instance Extraction}

\begin{figure}[!t]
    \centering
    \includegraphics[width=1.0\linewidth]{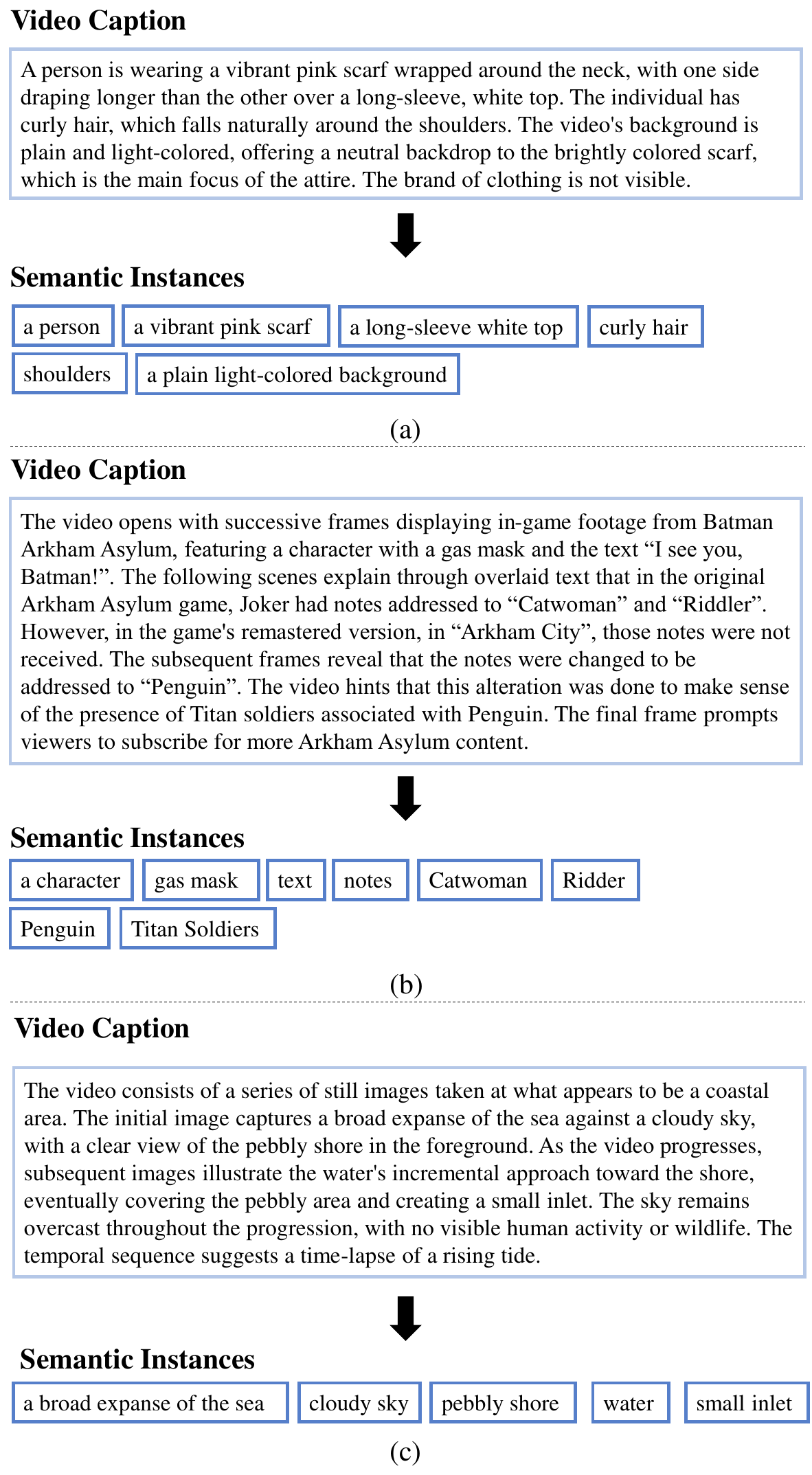}
    \caption{Examples of the semantic instance extraction process. Through our carefully designed prompts, the extracted instances do not undergo redundancy, while fully covers the major entities in the video caption.}
    \label{fig:supp-extract}
\end{figure}

In the Visual Concept Discriminator (VCD) module, to acquire distinct semantic concepts of training videos, we adopt GPT-4~\cite{achiam2023gpt} to extract words or phrases that correspond to specific entities in the video caption. In~\cref{fig:supp-prompt}, we show the prompts we use to guide GPT-4 to perform the extraction of semantic instances. Notably, we find that it is important to add the instruction on requiring GPT not to repeatedly draw the same instances that appear multiple times in the video caption (`Do not include repetitive objects' in~\cref{fig:supp-prompt}). Examples of the extracted instances in~\cref{fig:supp-extract}. We can see that our approach comprehensively draws out major instances in the caption, without containing repetitive items.

\section{More Ablations}

\noindent\textbf{Methods on Semantic Instance Extraction.} \ To verify the necessity of extracting non-overlapping instances in the semantic extraction process, we compare our `unoverlapped' extraction method with the simple approach of extracting all appeared instances (`overlapped'), even if there are repetitive items. From~\cref{tab:ablation-semantic}, we can see that although using our `unoverlapped' method results in a slight decrease in the average number of instances per video ($9.91$ v.s. $11.03$), our method consistently achieves better performance on all three benchmarks. These results validate the superiority of our semantic instance extraction method, while further consolidating the importance of relieving semantic redundancy in the learning process of visual tokens.

\noindent\textbf{Results on Varied Caption Quality.} \ In the VCD module, DisCo utilizes textual instances extracted from video captions. To explore the influence of caption quality (\eg, length, detailedness) on the final results, we utilize two sets of captions: (1) ShareGPT4o~\cite{zhang2024direct} which features high-quality dense captions, and (2) WebVid2M~\cite{bain2021frozen} which features short, brief captions. As shown in~\cref{tab:ablation-caption}, the two caption sources vary a lot in caption length and number of entities. ShareGPT4o captions contain an average of $9.96$ instances per sample, while WebVid2M captions could only yield $3.23$ instances per sample. Nevertheless, we observe that using both captions could result in a notable performance gain, with $1.9\%$ and $1.5\%$ improvement on MVBench, respectively. This highlighting DisCo's adaptability to different caption types. As the instance number in WebVid2M data is significantly less than ShareGPT4o data, for the training of WebVid2M captions, we decrease the number of tokens used in VCD module from $64$ to $24$, and decrease the number of token groups from $16$ to $6$, to reduce the proportion of unmatched visual tokens. 

\noindent\textbf{Ablations on Weights of Different Loss Functions.} \ Moreover, in~\cref{eq:stage1 loss}, the weights of each loss component are crucial hyperparamters that can largely affect the capability of the trained model. Therefore, in order to decide the best combinations of each hyperparameter, we carry out an ablation in~\cref{tab:ablation-loss}. Experimental results show that the model achieves an overall best performance when setting all weights $\lambda_{vsc}, \lambda_{vsm}, \lambda_{fsc}$ to $1.0$.

\noindent\textbf{Comparison with Other Token Compressing Methods.} \ In the area of MLLMs, there have been a series of token compression methods aiming at effectively representing visual features using fewer tokens, which share similarities with DisCo. In~\cref{tab:token compression}, we compare two related works, TokenPacker~\cite{li2024tokenpacker} and DeCo~\cite{yao2024deco}, with DisCo. As shown in~\cref{tab:token compression}, by using significantly fewer visual tokens (64 against 400/256), DisCo achieves comparable performance with TokenPacker and DeCo. At the same time, the training and inference time of DisCo largely outcompetes the other two methods, demonstrating the superiority of our visual encapsulation approach.

\begin{table}[]
    \centering
    \caption{Ablations on different methods of extracting semantic instances. EgoSchema is validated on \textit{subset}.}
    \resizebox{1.0\linewidth}{!}{
    \begin{tabular}{c|c|ccc}
    \toprule
    Method & Avg. Inst & MVBench & STAR & EgoSchema \\
    \midrule
    Overlapped & 11.03 & 67.8 & 76.0 & 71.6 \\
    \rowcolor{mygray}
    \textbf{Unoverlapped} & 9.96 & \textbf{68.2} & \textbf{77.7} & \textbf{72.2} \\
    \bottomrule
    \end{tabular}
    }
    \label{tab:ablation-semantic}
\end{table}

\begin{table}[]
\centering
\caption{Ablations on caption quality. We compare the results of adopting two set of captions: WebVid2M with short, sketchy captions and ShareGPT4o with long, detailed captions. `Avg words' and `Avg inst.' indicates the average number of words and extracted instances in each caption, respectively.}
\begin{adjustbox}{width=\linewidth,center}
\tablestyle{5pt}{1.0}
\begin{tabular}{l|cc|cc}
\toprule
\textbf{Method} & Avg words & Avg inst. & MVBench & STAR \\
\midrule
InternVideo2-HD & - & - & 66.3 & 75.7 \\
InternVideo2-HD+WebVid2M & 14.2 & 3.23 & 67.8 & 76.7 \\
InternVideo2-HD+ShareGPT4o & 109.3 & 9.96 & \textbf{68.2} & \textbf{77.7} \\
\bottomrule
\end{tabular}
\end{adjustbox}
\label{tab:ablation-caption}
\end{table}

\begin{table}[]
\centering
\caption{Ablations on the weights different components in the total training loss of DisCo. $\lambda_{vsc}$, $\lambda_{vsm}$, $\lambda_{ffa}$ indicates weights for the losses in Eq.7.}
\begin{adjustbox}{width=\linewidth,center}
\renewcommand{\arraystretch}{1.1}
\tablestyle{6pt}{1.0}
\begin{tabular}{ccc|ccc}
\toprule
$\lambda_{vsc}$ & $\lambda_{vsm}$ & $\lambda_{ffa}$ & MVBench & STAR & EgoSchema \\
\midrule
0.5 & 1.0 & 1.0 & 66.9 & 75.5 & 70.4 \\
2.0 & 1.0 & 1.0 & 68.0 & 76.4 & 71.1 \\
1.0 & 0.5 & 1.0 & 68.1 & 76.7 & 71.3 \\
1.0 & 2.0 & 1.0 & 67.4 & 76.4 & 69.8 \\
1.0 & 1.0 & 0.5 & 67.8 & \textbf{78.0} & 70.5 \\
1.0 & 1.0 & 2.0 & 66.5 & 75.4 & 69.7 \\
\midrule
\rowcolor{mygray}
\textbf{1.0} & \textbf{1.0} & \textbf{1.0} & \textbf{68.2} & 77.7 & \textbf{72.2} \\
\bottomrule
\end{tabular}
\end{adjustbox}
\label{tab:ablation-loss}
\end{table}

\begin{table}
    \centering
    \caption{Comparisons between DisCo and two other visual token compression methods in MLLMs, TokenPacker and DeCo. We compare the number of visual tokens, training time per step, inference time per instance, and the accuracy on MVBench.}
    \begin{adjustbox}{width=0.9\linewidth, center}
    \tablestyle{8pt}{1.0}
    \begin{tabular}{c|ccc}
    \toprule
        Model & DeCo & TokenPacker & \textbf{DisCo} \\
    \midrule
       Token No. & 400 & 256 & \textbf{64} \\
       Train time(s/step) & 6.9 & 6.4 & \textbf{4.6} \\
       Inference time(s/it) & 1.52 & 1.33 & \textbf{1.11} \\
       MVBench Acc. & 68.1 & 67.6 & \textbf{68.2} \\
    \bottomrule
    \end{tabular}
    \end{adjustbox}
    \label{tab:token compression}
\end{table}

{
    \small
    \bibliographystyle{ieeenat_fullname}
    \bibliography{main}
}

\end{document}